\documentclass[runningheads]{llncs}
\usepackage{graphicx} 
\usepackage{float} 
\usepackage{subfigure} 

\usepackage{latexsym}
\usepackage{graphicx}
\usepackage{url}
\usepackage{CJK}
\usepackage{soul}
\usepackage{url}
\usepackage{array}
\usepackage{xcolor}
\usepackage{adjustbox}
\usepackage{makecell}
\usepackage{hyperref}
\usepackage[small]{caption}
\usepackage{graphicx}
\usepackage{amsmath}
\usepackage{booktabs}
\usepackage{algorithm}
\usepackage{algorithmic}
\usepackage{float}

\usepackage{multirow}

\begin{document}

\title{Utilizing Longitudinal Chest X-Rays and Reports to Pre-Fill Radiology Reports\thanks{Corresponding author: zhiyong.lu@nih.gov}}
\titlerunning{Using Longitudinal Data to Pre-Fill Radiology Reports}



\author{
  Qingqing Zhu\inst{1}
  \and
  Tejas Sudharshan Mathai \inst{2}
  \and
  Pritam Mukherjee \inst{2}
  \and
  Yifan Peng \inst{3}
  \and
  Ronald M. Summers\inst{2}
  \and
  Zhiyong Lu\inst{1}
}

\authorrunning{Qingqing Zhu et al.}

\institute{National Center for Biotechnology Information, National Library of Medicine, National Institutes of
Health, Bethesda, MD, USA \and
Imaging Biomarkers and Computer-Aided Diagnosis Laboratory, Department of Radiology and Imaging Sciences, National Institutes of Health Clinical Center, Bethesda, MD, USA \and
Department of Population
Health Sciences, Weill Cornell Medicine,\\ New York, NY, USA}

%

\maketitle              

\begin{abstract}

Despite the reduction in turn-around times in radiology reporting with the use of speech recognition software, persistent communication errors can significantly impact the interpretation of radiology reports. Pre-filling a radiology report holds promise in mitigating reporting errors, and despite multiple efforts in literature to generate comprehensive medical reports, there lacks approaches that exploit the longitudinal nature of patient visit records in the MIMIC-CXR dataset. To address this gap, we propose to use longitudinal multi-modal data, i.e., previous patient visit CXR, current visit CXR, and the previous visit report, to pre-fill the ``findings'' section of the patient's current visit. We first gathered the longitudinal visit information for 26,625 patients from the MIMIC-CXR dataset, and created a new dataset called \textit{Longitudinal-MIMIC}. With this new dataset, a transformer-based model was trained to capture the multi-modal longitudinal information from patient visit records (CXR images + reports) via a cross-attention-based multi-modal fusion module and a hierarchical memory-driven decoder. In contrast to previous works that only uses current visit data as input to train a model, our work exploits the longitudinal information available to pre-fill the ``findings'' section of radiology reports. Experiments show that our approach outperforms several recent approaches.
Code will be published at \url{https://github.com/CelestialShine/Longitudinal-Chest-X-Ray}.

\keywords{Chest X-Rays \and Radiology reports \and Longitudinal data \and Report Pre-Filling \and Report Generation.}

\end{abstract}


\section{Introduction}

In current radiology practice, a signed report is often the primary form of communication, to communicate results of a radiological imaging exam between radiologist. Speech recognition software (SRS), which converts dictated words or sentences into text in a report, is widely used by radiologists. Despite SRS reducing the turn-around times for radiology reports, correcting any transcription errors in the report has been assumed by the radiologists themselves. But, persistent report communication errors due to SRS can significantly impact report interpretation, and also have dire consequences for radiologists in terms of medical malpractice \cite{Smith2001_legalRads}. These errors are most common for cross-sectional imaging exams (e.g., CT, MR) and chest radiography \cite{Ringler2017_errorsDictation}. Problems also arise when re-examining the results from external examinations and in interventional radiology procedural reports. Such errors are due to many factors, including SRS finding a nearest match for a dictated word, the lack of natural language processing (NLP) for real-time recognition and dictation conversion \cite{Ringler2017_errorsDictation}, and unnoticed typographical mistakes. To mitigate these errors, a promising alternative is to automate the pre-filling of a radiology report with salient information for a radiologist to review. This enables standardized reporting via structured reporting.

\begin{figure}[!htbp]
\centering
\includegraphics[width=\textwidth,height=0.35\textwidth]{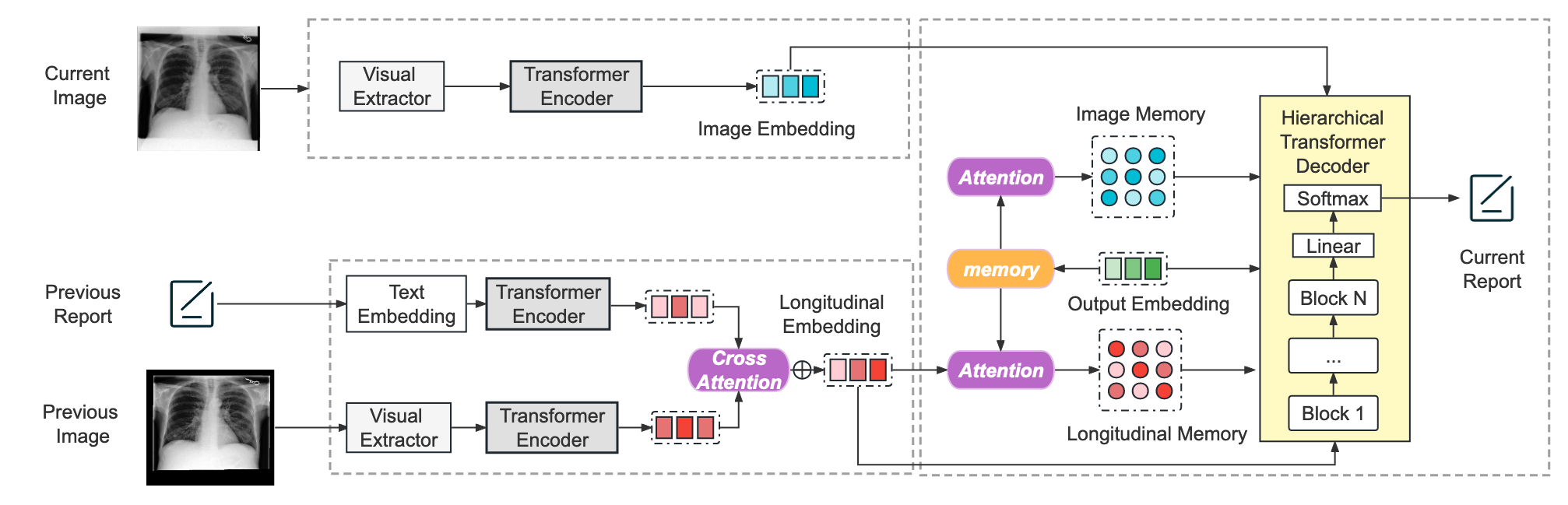}
\caption{Our proposed approach uses the CXR image and report from a previous patient visit and the current visit CXR image to pre-fill the ``findings'' section of the current visit report. The transformer-based model uses a cross-attention-based multi-modal fusion module and a hierarchical memory-driven decoder to generate the required text.}

\label{overall}
\end{figure}

A number of methods to generate radiology reports have been proposed previously, with significant focus on  CXR images \cite{shin2016learning,jing2018automatic,li2018hybrid,wang2018tienet,jing2020show,chen2020generating,chen2021cross,Wang2022,liu2021exploring}. Various attention mechanisms were proposed \cite{jing2018automatic,xuemultimodal,wang2018tienet} to drive the encoder and the decoder to emphasize more informative words in the report, or visual regions in the CXR, and improve generation accuracy. Other approaches \cite{chen2020generating,chen2021cross,Wang2022} effectively used Transformer-based models with memory matricies to store salient information for enhanced report generation quality. Despite these advances, there has been scarce research into harnessing \textit{the potential of longitudinal patient visits} for improved patient care. In practice, CXR images from multiple patient visits are usually examined simultaneously to find interval changes; e.g., a radiologist may compare a patient's current CXR to a previous CXR, and identify deterioration or improvement in the lungs for pneumonia. Reports from longitudinal visits contain valuable information regarding the patient's history, and harnessing the longitudinal multi-modal data is vital for the automated pre-filling of a comprehensive ``findings'' section in the report.

In this work, we propose to use longitudinal multi-modal data, i.e., previous visit CXR, current visit CXR, and previous visit report, to pre-fill the ``findings'' section of the patient's current visit report. To do so, we first gathered the longitudinal visit information for 26,625 patients from the MIMIC-CXR dataset\footnote{https://physionet.org/content/mimic-cxr-jpg/2.0.0/} and created a new dataset called \textit{Longitudinal-MIMIC}. Using this new dataset, we trained a transformer-based model containing a cross-attention-based multi-modal fusion module and a hierarchical memory-driven decoder to capture the features of longitudinal multi-modal data (CXR images + reports). In contrast to current approaches that only use the current visit data as input, our model exploits the longitudinal information available to pre-fill the ``findings'' section of reports with accurate content. Experiments conducted with the proposed dataset and model validate the utility of our proposed approach. Our main contribution in this work is training a transformer-based model that fully tackles the longitudinal multi-modal patient visit data to pre-fill the ``findings'' section of reports.

\section{Methods}

\noindent
\textbf{Dataset}. The construction of the Longitudinal-MIMIC dataset involved several steps, starting with the MIMIC-CXR dataset, which is a large publicly available dataset of 377,110 chest X-ray images corresponding to 227,835 radiographic reports from 65,379 patients \cite{johnson2019mimic}.
The first step in creating the Longitudinal-MIMIC dataset was to pre-process MIMIC-CXR to ensure consistency with prior works \cite{chen2021cross,chen2020generating}. Specifically, patient visits where the report did not contain a ``findings'' section were excluded. For each patient visit, there was at least one chest X-ray image (frontal, lateral or other view) and a corresponding medical report. In our work, we only generated pre-filled reports with the ``findings'' section.

\begin{table}[!htb]
\centering\fontsize{9}{12}\selectfont 
\setlength\aboverulesep{0pt}\setlength\belowrulesep{0pt} 
\setlength{\tabcolsep}{7pt} 
\setcellgapes{3pt}\makegapedcells 
\caption{A breakdown of the MIMIC-CXR dataset to show the number of patients with a specific number of visit records.}
\begin{adjustbox}{max width=\textwidth}
\begin{tabular}{@{} l|cccccc @{}} 
\toprule

\# visit records    &  1        & 2         & 3         & 4         & 5         & $>$5 \\
\# patients         & 33,922    & 10,490    & 5,079     & 3,021     & 1,968     & 6,067 \\

\bottomrule
\end{tabular}
\end{adjustbox}
\label{table_numPatientVisits}
\end{table}

Next, the pre-processed dataset was partitioned into training, validation, and test sets using the official split provided with the MIMIC-CXR dataset. 
Table \ref{table_numPatientVisits} shows that 26,625 patients in MIMIC-CXR had $\geq$ 2 visit records, providing a large cohort of patients with longitudinal study data that could be used for our goal of pre-filling radiology reports. 
For patients with $\geq$2  visits, consecutive pairs of visits were used to capture richer longitudinal information. 
The dataset was then arranged chronologically based on the ``StudyDate'' attribute present in the MIMIC-CXR dataset. ``StudyDate'' denotes a de-identified date linked to a radiographic analysis.  While the dates undergo anonymization, they maintain a consistent chronological sequence for each individual patient.

Following this, patients with $\geq$2 visit records were selected, resulting in 26,625 patients in the final \textit{Longitudinal-MIMIC} dataset with a total of 94,169 samples. Each sample used during model training consisted of the current visit CXR, current visit report, previous visit CXR, and the previous visit report. The final dataset was divided into training (26,156 patients and 92,374 samples), validation (203 patients and 737 samples), and test (266 patients and 2,058 samples) splits. We aimed to create the \textit{Longitudinal-MIMIC} dataset to enable the development and evaluation of models leveraging multi-modal data (CXR + reports) from longitudinal patient visits.

\medskip
\noindent
\textbf{Model Architecture}. Figure \ref{overall} shows the pipeline to generate a pre-filled ``findings'' section in the current visit report $R_{C}$, given the current visit CXR image $I_{C}$, previous visit CXR image $I_{P}$, and the previous visit report $R_{P}$. Mathematically, we can write:
$
p(R_{C} \mid I_{C},I_{P},R_{P})=\prod_{t=1} p\left({w}_t \mid {w}_1, \ldots, {w}_{t-1}, I_{C},I_{P},R_{P}\right),
$
where 
 $w_i$ is the $i$-th word in the current report.

\smallskip
\noindent
\textbf{Encoder}. Our model uses an \textit{Image Encoder} and a \textit{Text Encoder} to process the CXR images and text input separately. Both encoders were based on transformers. First, a pre-trained ResNet-101 \cite{simonyan2014very} extracted image features $F=[f_1 , \ldots , f_S ] $ from the CXR images, where $S$ is the number of patch features. They were then passed to the \textit{Image Encoder}, which consisted of a stack of blocks.  
The encoded output was a list of encoded hidden states $H=[h_1, \ldots , h_S]$. The CXR images from the previous and the current visits were encoded in the same manner, and denoted by ${H}^{I_{P}}$ and ${H}^{I_{C}}$ respectively.

The \textit{Text Encoder} encoded text information for language feature embedding using a previously published method \cite{DBLP:conf/nips/VaswaniSPUJGKP17}. First, the radiology report $R_{P}$ was tokenized into a sequence of $M$ tokens, and then transformed into vector representations $V = [{v}_1, \ldots, {v}_M]$ using a lookup table \cite{moon2021multi}. They were then fed to the \textit{text encoder}, which had the same architecture as the \textit{image encoder}, but with distinct network parameters. The final text feature embedding $H^{R_{P}}$ was defined as: $H^{R_{P}} = \theta^{E}_{R}(V)$, where $\theta^{E}_{R}$ refers to the parameters of the report text encoder.

\smallskip
\noindent
\textbf{Cross-Attention Fusion Module}. 
A multi-modal fusion module integrated longitudinal representations of images and texts using a cross-attention mechanism \cite{nagrani2021attention}, which was defined as: $
{H}^{I^{*}_{P}}=\operatorname{softmax}\left(\frac{q\left({H}^{I_{P}}\right) k\left({H}^{R_{P}}\right)^{\top}}{\sqrt{d_k}}\right) v\left({H}^{R_{P}}\right) 
$ and $
{H}^{R^{*}_{P}}={softmax}\left(\frac{q\left({H}^{R_{P}}\right) k\left({H}^{I_{P}}\right)^{\top}}{\sqrt{d_k}}\right) v\left({H}^{I_{P}}\right) \text {, }
$ where $q(\cdot), k(\cdot)$, and $v(\cdot)$ are linear transformation layers applied to features of proposals. $d_k$ is the number of attention heads for normalization. Finally, ${H}^{I^{*}_{P}}$ and ${H}^{R^{*}_{P}}$ were concatenated to obtain the multi-modal longitudinal representations $H^{L}$.

\smallskip
\noindent
\textbf{Hierarchical Decoder with Memory}. Our model's backbone decoder is a Transformer decoder with multiple blocks (The architecture of an example block  is shown in the supplementary material). The first block takes partial output embedding ${H}^{O}$ as input during training and a pre-determined starting symbol during testing. Subsequent blocks use the output from the previous block as input. To incorporate the encoded $H^{L}$ and ${H}^{I_{C}}$, we use a hierarchical structure for each block that divides it into two sub-blocks: $D^{I}$ and $D^{L}$.

Sub-block-1 uses ${H}^{I_{C}}$ and consists of a self-attention layer, an encoder-decoder attention layer, and feed-forward layers. It also employs residual connections and conditional layer normalization \cite{chen2020generating}. The encoder-decoder attention layer performs multi-head attention over ${H}^{I_{C}}$. It also uses a memory matrix $M$ to store output and important pattern information. The memory representations not only store the information of generated current reports over time in the decoder, but also the information across different encoders. Following \cite{chen2020generating}, we adopted a matrix $M$ to store the output over multiple generation steps and record important pattern information. Then we enhance $M$ by aligning it with ${H}^{I_{C}}$ to create an attention-aligned memory $M^{I_{C}}$ matrix. Different from \cite{chen2020generating}, we use $M^{I_{C}}$ while transforming the normalized data instead of $M$.  
The decoding process of sub-block-1 $D^{I}$ is formalized as:
$
H^{d e c, b,I}=D^{I}({H}^{O},{H}^{I_{C}},M^{I_{C}})
$, where $b$ stands for the block index. The output of sub-block 1 is combined with ${H}^{O}$ through a fusion layer:
$
{H}^{d e c, b}= (1-\beta){H}^{O}+\beta{H}^{d e c, b,I}.
$
$\beta$ is
a hyper-parameter to balance ${H}^{O}$ and ${H}^{d e c, b,I}$. In
our experiment, we set it to 0.2.  

The input to sub-block-2 $D^{L}$ is ${H}^{dec, b}$ . This structure is similar to sub-block-1, but interacts with $H^{L}$ instead of ${H}^{I_{C}}$. The output of this block is ${H}^{dec, b,L}$ and combined with ${H}^{dec, b,I}$ by adding them together. After fusing these embeddings and doing traditional layer normalization for them, we use these embeddings as the output of a block. The output of the previous block is used as the input of the next block.  After $N$ blocks, the final hidden states are obtained and used with a Linear and Softmax layer to get the target report probability distributions.

\section{Experiments and Results}

\noindent
\textbf{Baseline comparisons.} We compared our proposed method against prior image captioning and medical report generation works respectively. The same \textit{Longitudinal-MIMIC} dataset was used to train all baseline models, such as AoANet \cite{huang2019attention}, CNNTrans ~\cite{moon2021multi}, Transformer ~\cite{DBLP:conf/nips/VaswaniSPUJGKP17}, R2gen \cite{chen2020generating}, and R2CMN \cite{chen2021cross}. Implementation of these methods is detailed in the supplementary material.

\smallskip
\noindent
\textbf{Evaluation Metrics.} Conventional natural language generation (NLG) metrics, such as $BLEU$ \cite{DBLP:conf/acl/PapineniRWZ02}, $METEOR$ \cite{denkowski2014meteor}, and ${Rouge}_{L}$ \cite{lin2004rouge} were used to evaluate the utility of our approach against other baseline methods. Similar to prior work \cite{moon2021multi,chen2020generating}, the CheXpert labeler \cite{irvin2019chexpert} classified the predicted report for the presence of 14 disease conditions \footnote{No Finding, Enlarged Cardiomediastinum, Cardiomegaly, Lung Lesion, Airspace Opacity, Edema, Consolidation, Pneumonia, Atelectasis, Pneumothorax, Pleural Effusion, Pleural Other, Fracture and Support Devices} and compared them against the labels of the ground-truth report. Clinical Efficacy (CE) metrics, such as; accuracy, precision, recall, and F-1 score, were used to evaluate model performance.

\begin{table*}[!tb]\small 
	\centering

		\caption{Results of the NLG metrics (BLEU (BL), Meteor (M), Rouge $R_L$) and clinical efficacy (CE) metrics (Accuracy, Precision, Recall and F-1 score) on the \textit{Longitudinal-MIMIC} dataset. Best results are highlighted in bold.}
	\begin{tabular}{c|cccccc|cccc} \hline
	
		\multirow{2}{*}{Method} & \multicolumn{6}{c|}{ NLG metrics} & \multicolumn{4}{c}{ CE metrics}

		 \\ \cline{2-11}
		 & BL-1 & BL-2 & BL-3 & BL-4 & M & $R_L$ & A & P & R & F-1
 \\ \hline 
  	AoANet & 0.272 &

        0.168 & 0.112 & 0.080

       & 0.115

        & 0.249 & 0.798 & 0.437  & 0.249 & 0.317\\
		 CNN+Trans & 0.299 & 0.186 & 0.124 & 0.088 & 0.120 & 0.263 & 0.799 & 0.445 & 0.258 & 0.326
   \\
        Transformer &  0.294 & 0.178 & 0.119 & 0.085 & 0.123 & 0.256 & 0.811 & 0.500 & 0.320 & 0.390
		 \\
		R2gen & 0.302 &

      0.183

        &0.122

        & 0.087

        &0.124

        & 0.259 & 0.812 & 0.500 & 0.305 & 0.379

         \\ 

		R2CMN & 0.305 & 0.184 & 0.122 & 0.085 & 0.126 & 0.265 & 0.817 & 0.521 & 0.396 & 0.449\\
  \hline
  Ours & \textbf{0.343} & \textbf{0.210} & \textbf{0.140} & \textbf{0.099} & \textbf{0.137} & \textbf{0.271} & \textbf{0.823} & \textbf{0.538} & \textbf{0.434} & \textbf{0.480} 
  \\ \hline 
Baseline &  0.294 & 0.178 & 0.119 & 0.085 & 0.123 & 0.256 & 0.811 & 0.500 & 0.320 & 0.390
		 \\
  
		+ report & 0.333 & 0.201

      & 0.133 & 0.094 & 0.135 & 0.268 & 0.823 & 0.539 & 0.411 & 0.466
		 \\
   		+ image & 0.320 & 0.195

      & 0.130 & 0.092 & 0.130 & 0.268 & 0.817 & 0.522 & 0.34 & 0.412
     \\

		simple fusion & 0.317 & 0.193

      & 0.128 & 0.090 & 0.130 & 0.266 & 0.818 & 0.521 & 0.396 & 0.450
      \\
		
\hline

	\end{tabular}%

	\label{jieguo666}   
\end{table*}

\smallskip
\noindent
\textbf{Results.} Table \ref{jieguo666} shows the summary of the NLG metrics and CE metrics for the 14 disease observations for our proposed approach when compared against prior baseline approaches.  In particular, our model achieves the best performance over previous baselines across all NLG and CE metrics. 

Generic image captioning approaches like AoANet resulted in unsatisfactory performance on the \textit{Longitudinal-MIMIC} dataset as they failed to capture specific disease observations. Moreover, our approach outperforms previous report generation methods, R2Gen and R2CMN that also use memory-based models, due to the added longitudinal context arising from the use of longitudinal multi-modal study data (CXR images + reports). 
In our results, the BLEU scores show a substantial improvement, particularly in BLEU-4, where we achieve a 1.4\% increase compared to the previous method R2CMN. BLEU scores measure how many continuous sequences of words appear in predicted reports, while ${Rouge}_{L}$ evaluates the fluency and sufficiency of predicted reports. The highest ${Rouge}_{L}$ score demonstrates the ability of our approach to generate accurate reports, rather than meaningless word combinations. We also use METEOR for evaluation, taking into account the precision, recall, and alignment of words and phrases in generated reports and the ground truth. Our METEOR score shows a 1.1\% improvement over the previous outstanding method, which further solidifies the effectiveness of our approach. Meanwhile, our model exhibits a significant improvement in clinical efficacy metrics compared to other baselines. Notably, F1 is the most important metric, as it provides a balanced measure of both precision and recall. Our approach outperforms the best-performing method by 3.1\% in terms of F1 score. These observations are particularly significant, as higher NLG scores do not necessarily correspond to higher clinical scores \cite{chen2020generating}, confirming the effectiveness of our proposed method.

\smallskip
\noindent
\textbf{Effect of Model Components}.
We also studied the contribution of different model components and detail results in Table \ref{jieguo666}. The \textit{Baseline} experiment refers to a basic Transformer model trained to generate a pre-filled report given a chest CXR image without any additional longitudinal information. The NLG and CE metrics are poor for the vanilla transformer compared to our proposed approach. We also analyze the contributions of the previous chest CXR image \textit{+ image} and previous visit report \textit{+ report} when added to the model separately. These two experiments included memory-enhanced conditional normalization. We observed that with each added feature enhanced the pre-filled report quality compared to the baseline, but the previous visit report had a higher impact than the previous CXR image. We hypothesize that the previous visit reports contain more text that can be directly transferred to the current visit reports. In our \textit{simple fusion} experiment, we removed the cross-attention module and concatenated the encoded embeddings of the previous CXR image and previous visit report as one longitudinal embedding, while retaining the rest of the model. We saw a performance drop compared to our approach on our dataset, and also noticed that the results were worse than using the images or reports alone. These experiments demonstrate the utility of the cross-attention module in our proposed work.


\begin{figure}[!htb]
\centering
\includegraphics[width=0.95\textwidth]{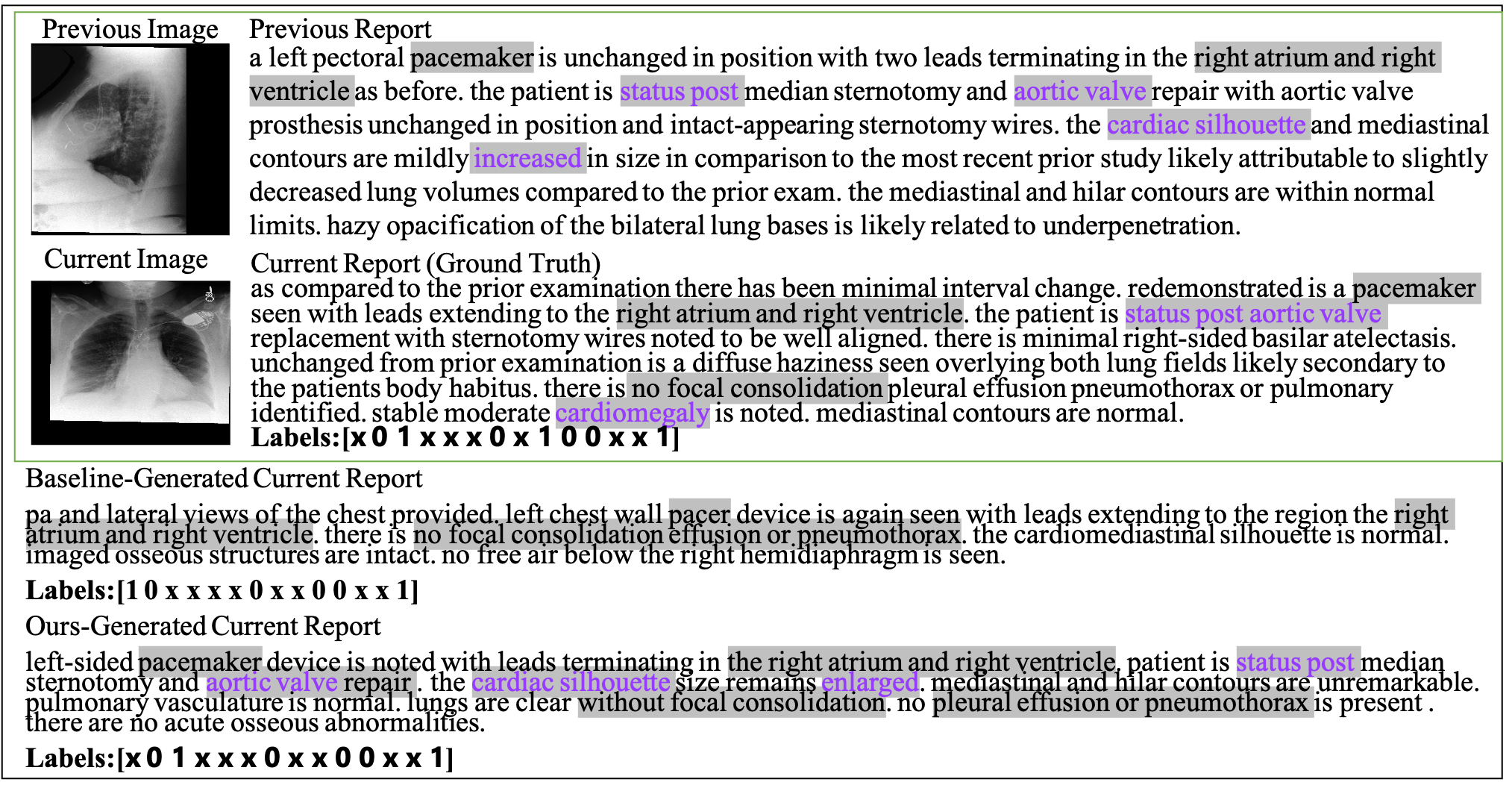}
\includegraphics[width=0.95\textwidth]{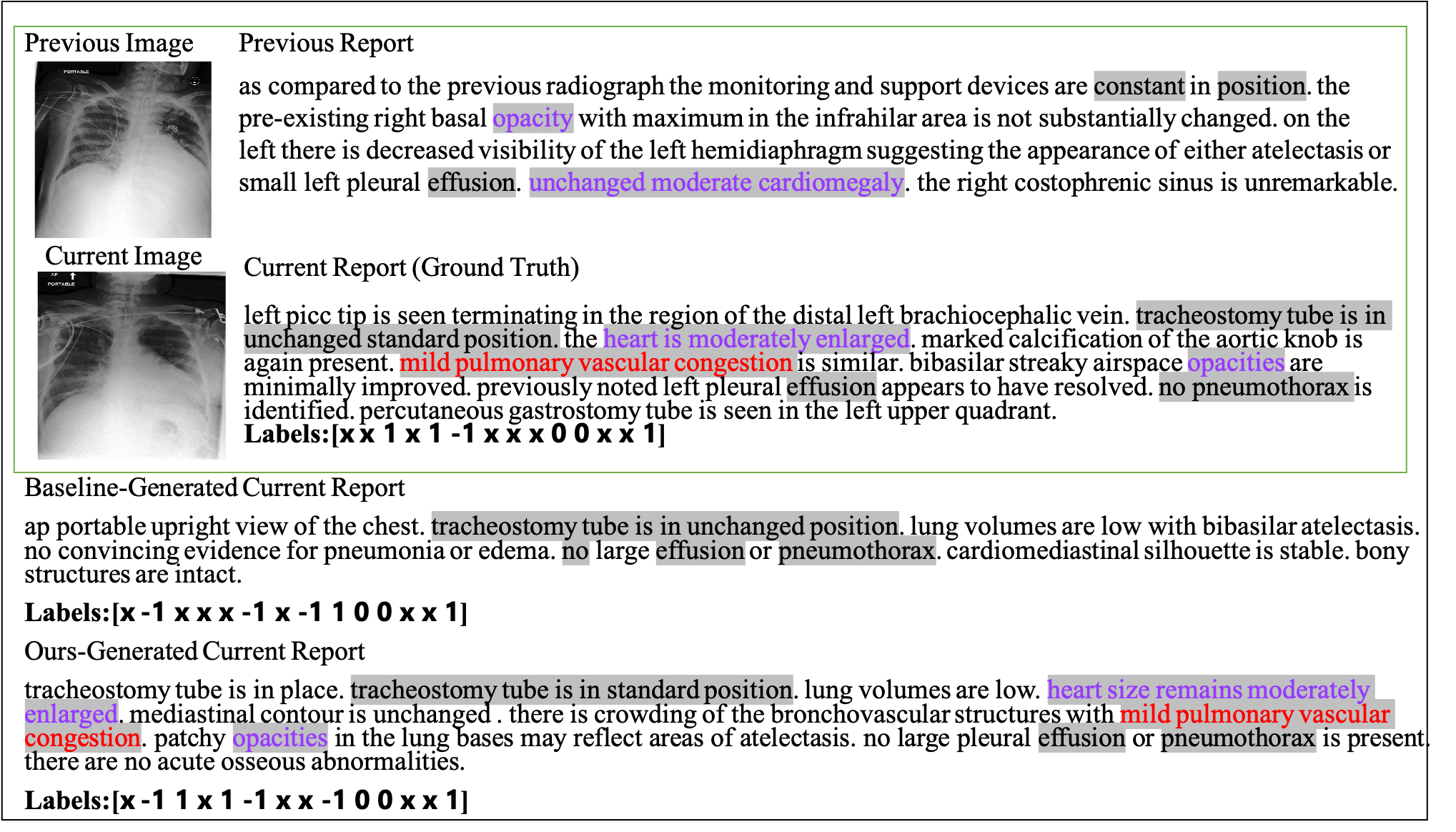}
\caption{Two examples of  pre-filled ``findings'' sections of  reports. Gray highlighted text indicates the same words or words with similar meaning that appear in the current reports and other reports. Purple highlighted text represents similar words in the current visit report generated by our approach, previous visit reports, and groundtruth current visit report.
The red highlighted text indicates similar words that only exist in the report generated by our approach and the current ground truth report.
R2Gen was the baseline method that generated the report. The ``Labels'' array shows the CheXpert classification of 14 disease observations (see text for details) as positive (1), negative (-1), uncertain (0) or unmentioned ($\times$).}

\label{exap}
\end{figure}
\section{Discussion and Conclusion}

\textbf{Case Study}. We also ran a qualitative evaluation of our proposed approach on two cases as seen in Fig. \ref{exap}. In these cases, we compare our generated report with the report generated by the R2Gen. 
In the first case, certain highlighted words in purple, such as ``status post'', ``aortic valve'' and ``cardiac silhouette  in the predicted current visit report are also seen in the previous visit report. 
The CheXpert classified ``Labels'' also show the pre-filled ``findings'' generated is highly consistent with the ground truth report in contrast to the R2Gen model. For example, the ``cardiac silhouette enlarged'' was not generated by the R2Gen model, but our prediction contains them and is consistent with the word ``cardiomegaly'' in the ground truth report. 
In the second case, our generated report is also superior. Not only does our report generate more of the same content as the ground truth, but the positive diagnosis labels classified by CheXpert in our report are completely consistent with those in the ground truth. 
We also provide more cases in the supplementary material.

\smallskip
\noindent
\textbf{Error Analysis}. 
To analyze errors from our model, we examine generated reports alongside ground truths and longitudinal information.
It is found that the label accuracy of the observations in the generated reports is greatly affected by the previous information. 
For example, as time changes, for the same observation ``pneumothorax'', the label can change from ``positive'' to ``negative''. And such changing examples are more difficult to generate accurately. 
According to our statistics, on the one hand, when the label results of current  and previous report are the same, 88.96\% percent of the generated results match them.  On the other hand, despite mentioning the same observations, when the labels of current  and previous report  are different, there is an 84.42\% probability of generated results being incorrect. Thus how to track and generate the label accurately of these examples is a  possible future work to improve the generated radiology reports. 
One possible way to address this issue is to use active learning \cite{settles2012active} or curriculum learning \cite{bengio2009curriculum} methods to differentiate different types of samples and better train the machine learning models. 

\smallskip
\noindent
\textbf{Conclusion.} In this paper, we propose to pre-fill the ``findings'' section of chest X-Ray radiology reports by considering the longitudinal multi-modal (CXR images + reports) information available in the MIMIC-CXR dataset. We gathered 26,625 patients with multiple visits to constitute the new \textit{Longitudinal-MIMIC} dataset, and proposed a model to fuse encoded embeddings of multi-modal data along with a hierarchical memory-driven decoder. The model generated a pre-filled ``findings'' section of the report, and we evaluated the generated results against prior image captioning and medical report generation works. Our model yielded a $\geq$ 3\% improvement in terms of the clinical efficacy F-1 score on the \textit{Longitudinal-MIMIC} dataset. Moreover, experiments that evaluated the utility of different components of our model proved its effectiveness for the task of pre-filling the ``findings'' section of the report.  

\section{Acknowledgements}
   This research was supported by the Intramural Research Program of the National Library of Medicine and Clinical Center at the NIH. The authors thank to Qingyu Chen and Xiuying Chen for their time and effort in providing thoughtful comments and suggestions to revise this paper. This work was also supported by the National Institutes of Health under Award No. 4R00LM013001 (Peng), NSF CAREER Award No. 2145640 (Peng), and Amazon Research Award (Peng).

\clearpage
\bibliographystyle{unsrt} 
\bibliography{paper843}






\end{document}